\documentclass[10pt,twocolumn,letterpaper]{article}

\usepackage[pagenumbers]{cvpr} 

\usepackage{graphicx}
\usepackage{amsmath}
\usepackage{amssymb}
\usepackage{booktabs}
\usepackage{dblfloatfix}

\usepackage[pagebackref,breaklinks,colorlinks]{hyperref}

\usepackage[capitalize]{cleveref}
\crefname{section}{Sec.}{Secs.}
\Crefname{section}{Section}{Sections}
\Crefname{table}{Table}{Tables}
\crefname{table}{Tab.}{Tabs.}

\def\cvprPaperID{54} 
\def\confName{CVPR}
\def\confYear{2023}

\begin{document}

\title{Speed Is All You Need: On-Device Acceleration of Large Diffusion Models via GPU-Aware Optimizations}

\author{
    Yu-Hui Chen\thanks{These authors contributed equally to this work.}, Raman Sarokin\footnotemark[1], Juhyun Lee, Jiuqiang Tang,\\
    Chuo-Ling Chang, Andrei Kulik, Matthias Grundmann\\
    Google LLC\\
    1600 Amphitheatre Parkway Mountain View, CA 94043\\
    {\tt\small {yuhuic,sorokin,impjdi,jqtang,chuoling,akulik,grundman}@google.com}
}
\maketitle

\begin{abstract}
The rapid development and application of foundation models have revolutionized the field of artificial intelligence. Large diffusion models have gained significant attention for their ability to generate photorealistic images and support various tasks. On-device deployment of these models provides benefits such as lower server costs, offline functionality, and improved user privacy. However, common large diffusion models have over 1 billion parameters and pose challenges due to restricted computational and memory resources on devices. We present a series of implementation optimizations for large diffusion models that achieve the fastest reported inference latency to-date(under $12$ seconds for Stable Diffusion 1.4 without \textsc{int8} quantization 
for a $512\times512$ image with $20$ iterations) on GPU-equipped mobile devices. These enhancements broaden the applicability of generative AI and improve the overall user experience across a wide range of devices.
\end{abstract}

\section{Introduction}
\label{sec:intro}


The success of foundation models attributed to the advent and refinement of Generative Adversarial Networks (GANs)~\cite{GAN,DCGAN,StyleGAN} and Variational Autoencoders (VAEs)~\cite{VAE}, which have emerged as the leading approaches for creating high-quality images. Recently, diffusion-based generative models~\cite{DDPM,LatentStableDiffusion} that rely on the process of reverse diffusion to reconstruct images from noise have gained prominence in the image generation realm. One such model is Stable Diffusion~\cite{LatentStableDiffusion} that has attracted people's attention for its ability to generate photo-realistic images and its flexibility to support various tasks, including image-editing, in-painting, text-to-image generation, \etc. Its availability have made it a frequent inference benchmark target~\cite{AppleStable,QualcommStable}.

A crucial consideration for incorporating large diffusion models into any application is the choice of where the models will be executed. On-device generative AI deployment offers advantages such as reduced server costs, improved scalability, offline functionality, and enhanced user privacy due to local data processing.


Nonetheless, deploying large diffusion models like Stable Diffusion 1.4, which encompasses over 1 billion parameters, to on-device presents difficulties due to restricted computational and memory resources. In the absence of meticulous design and implementation, running these models on-device can lead to increased latency stemming from the iterative denoising process and excessive memory consumption. Although several successful endeavors~\cite{AppleStable,QualcommStable} have been made to deploy Stable Diffusion to on-device, these efforts are often limited to specific devices or chipsets and leave ample room for improving inference latency. Overcoming this constraint will further broaden the applicability of Generative AI and improve the overall user experience across a wider array of devices.

In this paper, we introduce implementation enhancements for large diffusion models, achieving state-of-the-art inference latency performance of executing Stable Diffusion 1.4 on GPU-powered devices (11.5 seconds on Samsung S23 Ultra to generate a $512\times512$ image with $20$ iterations). 

\section{Related Work}
\label{related_work}

\textbf{Image generation} has garnered significant research interest, particularly in recent years, with the advent of GANs~\cite{GAN}. GANs consist of two neural networks, a generator and a discriminator, that compete with each other to create realistic images~\cite{DCGAN,StyleGAN}. While GANs demonstrate the capability to generate high resolution images with good perceptual quality~\cite{LargeScaleGAN}, it suffers from the difficulty to train effectively~\cite{WassersteinGAN}. VAEs~\cite{VAE} emerged as a popular generative model, utilizing probabilistic graphical models to generate images through latent space representation, enabling efficient synthesis but with lower sample quality than GANs.


Denoising Diffusion Probabilistic Model (DDPM)~\cite{DDPM} marked a significant milestone in the development of diffusion-based generative models. DDPM demonstrated the potential of these models to generate high-quality images through a series of iterative noise-removal steps. More recently, Stable Diffusion~\cite{LatentStableDiffusion} emerged as a prominent diffusion-based model, attracting interest due to its capability to generate photorealistic images. Its open accessibility has further encouraged the community to extend and build upon the model~\cite{ControlNet,DreamBooth}.

\textbf{On-device model inference acceleration} has gained increasing attention as it offers several advantages over traditional server-based approaches, including reduced latency, enhanced privacy, and improved scalability. The softmax operation, common in deep learning, has prompted optimization efforts due to its complexity, leading to various acceleration approaches~\cite{PseudoSoftmax,LogSoftmax,HaSoftmax}. Winograd Convolution~\cite{WinogradConv} was introduced to optimize convolutional computation by reducing multiplications, resulting in faster processing and lower power consumption, particularly on GPUs.



The Transformer architecture~\cite{Attention} has been quite effective and popular, triggering active research on accelerating the attention mechanism. Reformer~\cite{Reformer} aims to minimize computational expense using sparse approximation, while other works~\cite{Performer,Longformer} employ low-rank or a mixture approximation methods. In contrast, FlashAttention~\cite{FlashAttention} is an exact attention algorithm that considers hardware settings for improved performance.

\section{GPU-Aware Optimizations}
\label{method}

\begin{figure}[t]
\centering
\includegraphics[width=180pt]{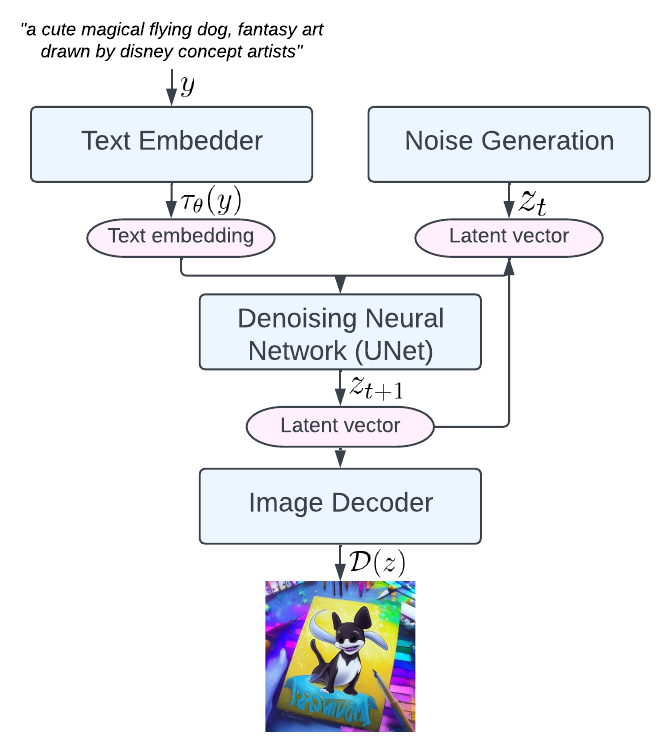}
\caption{Schematic representation of the primary components in Stable Diffusion and their interactions.}
\label{fig:stable_diffusion}
\end{figure}

Our primary emphasis is on the task of generating images from textual descriptions using large diffusion models. While the following description is discussing our proposed optimizations for the specific architecture of Stable Diffusion, those optimizations will readily apply to other large diffusion models. When performing inference with a text prompt, the process involves guiding the reverse diffusion process using additional conditioning based on the desired textual description. Specifically, the main components of Stable Diffusion include: text embedder, noise generation, denoising neural network, and image decoder, as shown in Figure~\ref{fig:stable_diffusion}.

\begin{itemize}
  \item \textbf{Text Embedder}: Utilizes the CLIP model~\cite{Unet} to encode the text prompt, $y$, resulting in a high-dimensional embedding vector, $\tau_\theta(y)$, that encapsulates the semantics of the input prompt. This embedding is employed as input to the denoising neural network, furnishing conditional guidance for the reverse diffusion process.
  \item \textbf{Noise Generation}: Supplies the random noise in the latent space, $z$, which functions as the initiation point for the reverse diffusion process.
  \item \textbf{Denoising Neural Network}: The network is designed to approximate conditional distributions of the form $p(z|y)$, utilizing a conditional denoising autoencoder, $\epsilon_\theta(z_t, t, \tau_\theta(y))$. Each iteration $t$ employs the UNet architecture~\cite{Unet}. The cross-attention mechanism~\cite{Attention} is adopted to operate on the latent space and the text embedding vector, predicting a denoised version of the input $z_t$ during the iterative procedure.
  \item \textbf{Image Decoder}: The reverse diffusion process is conducted in the latent space: $z=\mathcal{E}(x)\in\mathcal{R}^{h\times w\times c}$, where $x\in\mathcal{R}^{H\times W\times 3}$ represents the RGB image space. Once the process is completed, the image decoder $\mathcal{D}$ is used to reconstructs the RGB image from the latent vector: $\hat{x}=\mathcal{D}(\hat{z})$.
\end{itemize}


In this section, we introduce a set of optimization techniques aimed at enhancing the performance of running the diffusion-based model.

\subsection{Specialized Kernels: Group Norm and GELU}
\label{sec:special_kernels}

Group normalization (GN)~\cite{GroupNorm} is implemented throughout the UNet architecture as described in~\cite{DDPM}. This normalization technique works by dividing the channels of a feature map into smaller groups and normalizing each group independently, making GN less dependent on the batch size and more suitable for a wide range of batch sizes and network architectures. Each feature value $x_i$ is normalized by the group mean $\mu_g$ and variance $\sigma_g$ of the group it belongs to using Eq.~\ref{eq:gn}.

\begin{equation}
\label{eq:gn}
\hat{x_i}=\frac{1}{\sigma_g}(x_i-\mu_g)
\end{equation}

Rather than executing the aforementioned operations, which involves ``reshape'', ``mean'', ``variance'', ``normalize'', sequentially, we devise a unique kernel in the form of a GPU shader that executes all of them in a single GPU command without any intermediate tensors.

The Gaussian Error Linear Unit (GELU)~\cite{Gelu} serves as the prevalent activation function in the model, containing numerous numerical computations such as multiplications, addition, and the Gaussian error function, as shown in Eq.~\ref{eq:gelu}. We implemented a dedicated shader to consolidate these numerical computations and its accompanied split and multiplication ops, enabling their execution in a single draw call.

\begin{equation}
\label{eq:gelu}
\mathrm{GELU}(x) = \frac{x}{2}[1 + \mathrm{erf}(x/\sqrt{2})]
\end{equation}

\subsection{Enhancing Attention Module Efficiency}
\label{sec:attention_opt}

The text/image transformer within Stable Diffusion facilitates modeling the conditional distribution $P(z|\tau_\theta(y))$, which is crucial for the text-to-image generation task. Nonetheless, the self/cross-attention mechanism encounters difficulties with long sequences, owing to their quadratic time and memory complexity. In this section, we introduce two possible optimizations designed to alleviate these computational bottlenecks.

\subsubsection{Partially Fused Softmax}

The attention computation is adopted in the intermediate layers of the UNet

\begin{equation}
\label{eq:attention}
\mathrm{Attention}(Q,K,V) = \mathrm{softmax}(\frac{QK^T}{\sqrt{d}})\times V
\end{equation}

\noindent where $Q\in\mathcal{R}^{N\times d}; K, V\in \mathcal{R}^{M\times d}, $ corresponding to the query, key and value matrices as described in the original paper~\cite{LatentStableDiffusion} and typically $N, M$ are much larger than $d$.

The softmax operation performed on the matrix $A=\frac{QK^T}{\sqrt{d}}\in\mathcal{R}^{N\times M}$ can be partitioned into two steps: 1) reduction operations; 2) element-wise operations. The reduction operations refer to the calculation of the maximum values of each row in $A$ and its modified exponential sum $S$, as in Eq.\ref{eq:reduction_ops}. Subsequently, the element-wise operation is employed to normalize the values in $A$ utilizing the vectors $L$ and $S$.

\begin{equation}
\label{eq:reduction_ops}
L=[\max_{j} a_{ij}], S=[\sum_{j} \exp{(a_{ij}-\max_{k}a_{ik}})]\in\mathcal{R}^{N}
\end{equation}

\begin{figure}[t]
\centering
\includegraphics[width=230pt]{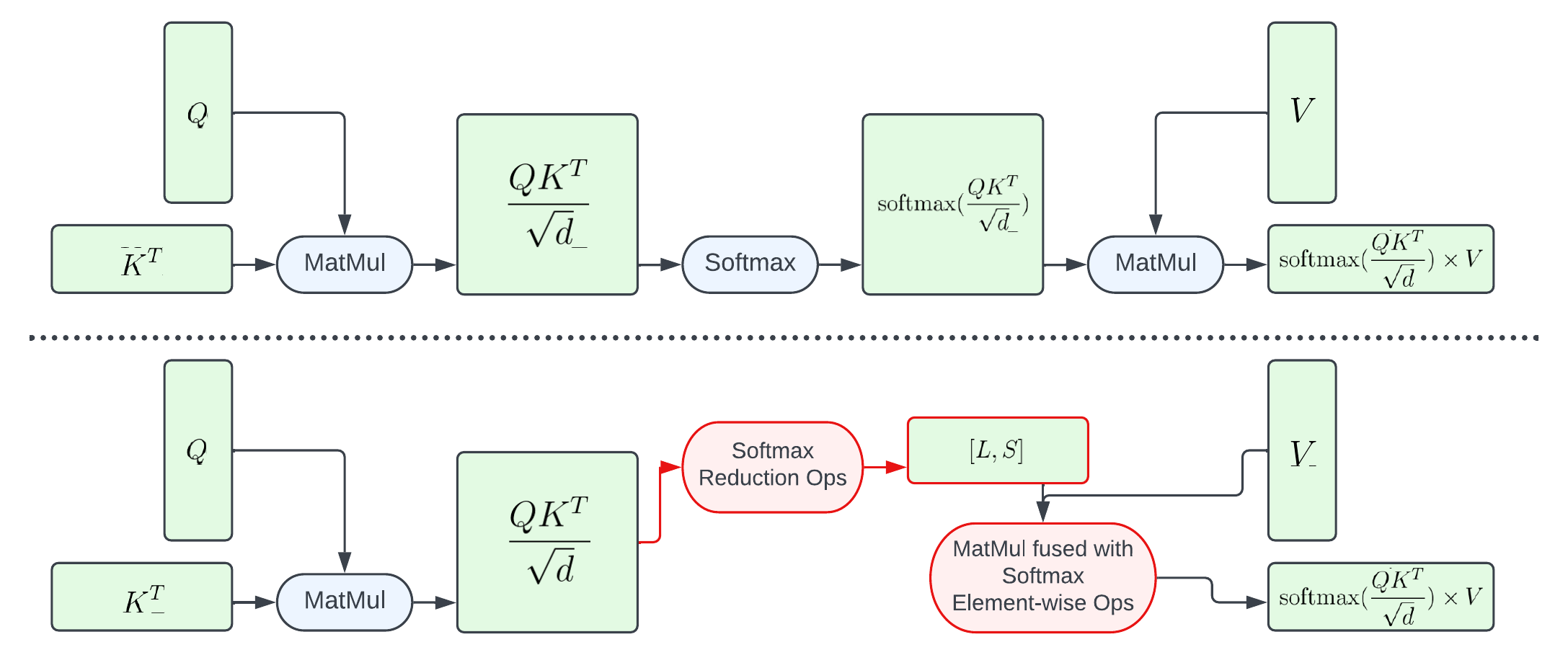}
\caption{Our optimized softmax implementation within the attention block. The upper diagram depicts the original implementation applying softmax directly to the matrix $\frac{QK^T}{\sqrt{d}}$; the lower diagram demonstrates the modified modules (highlighted in red).}
\label{fig:attention_opt}
\end{figure}

In order to avoid executing the whole softmax computation on the large matrix $A$, we implemented a GPU shader for the reduction operations to compute the $L$ and $S$ vectors, resulting in a tensor of size $N\times 2$. The element-wise softmax computation is then fused with the following matrix multiplication involving matrix $V$. This approach substantially reduces the memory footprint of the intermediate tensors and overall latency (Figure~\ref{fig:attention_opt}).

It is crucial to highlight that the parallelism of the computation mapping from $A$ to $L, S$ is limited, as the number of elements in the resulting tensors is considerably smaller than those in the input tensor $A$. To enhance parallelism and further decrease latency, we partition the reduction operations into multiple stages by grouping the elements in $A$ into blocks. The calculations are performed on each block, which are then reduced to the final result. By employing meticulously designed threading and memory cache management, this multi-stage implementation can be finished with a single GPU command and leads to additional latency reduction.

\subsubsection{FlashAttention}

There are many approximate attention methods~\cite{Reformer,Performer,Longformer} that attempt to improve the attention module latency by sacrificing model quality for reduced computational complexity. In contrast, FlashAttention~\cite{FlashAttention} is an IO-aware, exact attention algorithm that utilizes tiling to minimize memory reads/writes between GPU high bandwidth memory (HBM) and on-chip SRAM. This approach results in fewer HBM accesses than standard attention, making it optimal for a range of SRAM sizes and improving overall efficiency.

Although FlashAttention seeks to enhance latency and reduce global memory read/write, its kernel is highly register-intensive. Consequently, we selectively employ this technique for attention matrices with dimension $d=40$ on Adreno and Apple GPUs. In other cases, the partially fused softmax described in the previous section is utilized.

\begin{table*}[hbt!]
\centering
\begin{tabular}{|c|c|c|c|c|c|c|}
\hline
 & \multicolumn{3}{c|}{Samsung S23 Ultra} & \multicolumn{3}{c|}{iPhone 14 Pro Max} \\
\cline{2-7}
                & Latency  & Tensor  & Weight  & Latency  & Tensor  & Weight   \\
\hline
Baseline        & 1098     & 105     & 1640    & 1554     & 105     & 1640   \\
\hline
+Opt. Softmax   & 908      & 105     & 1640    & 1475     & 105     & 1640   \\
\hline
+S-GN/GELU      & 755      & 85      & 1640    & 1382     & 85      & 1640   \\
\hline
+FlashAttn.     & 660      & 80      & 1639    & 1309     & 80      & 1639   \\
\hline
+Winograd(\textbf{All})   & \textbf{525}      & 84      & 2093    & \textbf{1043}     & 84      & 2093   \\
\hline
\end{tabular}
\caption{Benchmark results for different devices with each optimization enabled. The numbers in the ``Latency'' column are in \emph{milliseconds} and the numbers in the ``Tensor'' and ``Weight'' columns are in \emph{megabytes}, representing the memory usage at runtime for intermediate tensors and at initialization time for model weights/biases. See text for details.}
\label{table:benchmark}
\end{table*}

\subsection{Winograd Convolution}
\label{sec:winograd}

Winograd convolution transforms the convolution operation into a series of matrix multiplications. The key insight is that by carefully choosing the transformation matrices, many of the required multiplications can be reduced, leading to a more efficient computation. However, it also introduces increased memory consumption and numerical errors, particularly when using larger tile sizes.

The backbone of Stable Diffusion relies heavily on $3\times3$ convolution layers, especially in the image decoder, where they comprise over $90\%$ of the layers. As a result, we delved into an analysis, as depicted in ~\cref{table:winograd}, to explore the potential benefits of employing Winograd with varying tile sizes on the $3\times3$ kernel convolutions. Our findings led us to select a $4\times 4$ tile size, as it provides an optimal balance between computational efficiency and memory utilization. Additionally, we strategically applied Winograd based on heuristic rules, only where it would produce profitable results, to further maximize its efficacy.

\begin{table}[htbp]
\centering
\begin{tabular}{c|ccc}
\toprule
Tile size & Flops Saving & Tensor & Weight \\
\midrule
$2\times2$ & 2.25x & 4x & 1.77x \\
$4\times4$ & 4x & 2.25x & 4x \\
$6\times6$ & 5.06x & 1.8x & 7.12x \\
$8\times8$ & 5.76x & 1.56x & 11.12x \\
\bottomrule
\end{tabular}
\caption{The impact of employing Winograd convolution with varying tile sizes in comparison with the standard $3\times 3$ convolution implementation. ``Flops Saving'' indicates the factor by which the algorithm reduces the number of flops. The columns ``Tensor'' and ``Weight'' represent the multiple by which the memory for hosting intermediate tensors and model weights increases, respectively.}
\label{table:winograd}
\end{table}

\section{Experiment}
\label{experiment}

To evaluate the improvement, we carried out a set of benchmarks on various devices, as displayed in ~\cref{table:benchmark}. We chose the following target devices (and their GPU chip): Samsung S23 Ultra (Adreno 740), and iPhone 14 Pro Max (A16). As the denoising neural network, UNet, is the most computationally demanding component, we provide the latency figures for executing UNet for a single iteration with $512\times512$ image resolution, measured in \emph{milliseconds}. Additionally, we document the memory usage generated during runtime for the intermediate tensors in the ``Tensor'' column and the memory allocated for holding the model weights in the ``Weight'' column, both quantified in \emph{megabytes}. Note that the memory manager~\cite{MemManager} optimizes the memory footprint by reusing the buffers for intermediate tensors.

The table's first row displays the result that follows the implementation in the public Github repository~\cite{CompVis} using our internal OpenCL kernels without any optimizations. Rows 2 through 5 sequentially enable each optimization individually: ``Opt. Softmax'' refers to the partially fused softmax and optimized softmax reduction step (subsection~\ref{sec:attention_opt}; ``S-GN/GELU'' relates to the specialized kernels for Group Normalization and GELU (subsection~\ref{sec:special_kernels}; ``FlashAttn.'' pertains the FlashAttention implementation (subsection~\ref{sec:attention_opt}); ``Winograd(All)'' employs the Winograd convolution (subsection~\ref{sec:winograd}). Note that each row includes all optimizations of its preceding rows, making the last row our final optimization numbers. 

It is evident that latency decreases incrementally as each optimization is activated. Notable overall latency reductions in comparison to the baseline are observed on both devices: Samsung S23 Ultra ($\mathbf{-52.2\%}$); iPhone 14 Pro Max ($\mathbf{-32.9\%}$). Additionally, we assessed the end-to-end latency (from text input to decoded image output) on the Samsung S23 Ultra for $20$ denoising iterations steps to generate a $512\times512$ pixel image, achieving a state-of-the-art result under $12$ seconds.

\section{Conclusion}
\label{conclusion}

In this paper, we presented a set of optimizations that collectively attain groundbreaking latency figures for executing large diffusion models on various devices. These improvements expand the model's versatility and enhance the overall user experience across an extensive array of devices.

{\small
\bibliographystyle{ieee_fullname}
\bibliography{egbib}

\begin{thebibliography}{10}\itemsep=-1pt

\bibitem{WassersteinGAN}
Martin Arjovsky, Soumith Chintala, and Léon Bottou.
\newblock Wasserstein gan, 2017.

\bibitem{Longformer}
Iz Beltagy, Matthew~E. Peters, and Arman Cohan.
\newblock Longformer: The long-document transformer.
\newblock {\em CoRR}, abs/2004.05150, 2020.

\bibitem{LargeScaleGAN}
Andrew Brock, Jeff Donahue, and Karen Simonyan.
\newblock Large scale gan training for high fidelity natural image synthesis,
  2019.

\bibitem{PseudoSoftmax}
Gian~Carlo Cardarilli, Luca Di~Nunzio, Rocco Fazzolari, Daniele Giardino,
  Alberto Nannarelli, Marco Re, and Sergio Span{\`o}.
\newblock A pseudo-softmax function for hardware-based high speed image
  classification.
\newblock {\em Sci. Rep.}, 11(1):15307, July 2021.

\bibitem{Performer}
Krzysztof Choromanski, Valerii Likhosherstov, David Dohan, Xingyou Song,
  Andreea Gane, Tam{\'{a}}s Sarl{\'{o}}s, Peter Hawkins, Jared Davis, Afroz
  Mohiuddin, Lukasz Kaiser, David Belanger, Lucy~J. Colwell, and Adrian Weller.
\newblock Rethinking attention with performers.
\newblock {\em CoRR}, abs/2009.14794, 2020.

\bibitem{FlashAttention}
Tri Dao, Daniel~Y. Fu, Stefano Ermon, Atri Rudra, and Christopher Ré.
\newblock Flashattention: Fast and memory-efficient exact attention with
  io-awareness, 2022.

\bibitem{HaSoftmax}
Xue Geng, Jie Lin, Bin Zhao, Anmin Kong, Mohamed M.~Sabry Aly, and Vijay
  Chandrasekhar.
\newblock Hardware-aware softmax approximation for deep neural networks.
\newblock In {\em Computer Vision – ACCV 2018: 14th Asian Conference on
  Computer Vision, Perth, Australia, December 2–6, 2018, Revised Selected
  Papers, Part IV}, page 107–122, Berlin, Heidelberg, 2018. Springer-Verlag.

\bibitem{GAN}
Ian~J. Goodfellow, Jean Pouget-Abadie, Mehdi Mirza, Bing Xu, David
  Warde-Farley, Sherjil Ozair, Aaron Courville, and Yoshua Bengio.
\newblock Generative adversarial networks, 2014.

\bibitem{Gelu}
Dan Hendrycks and Kevin Gimpel.
\newblock {Gaussian Error Linear Units (GELUs)}.
\newblock 2016.

\bibitem{DDPM}
Jonathan Ho, Ajay Jain, and Pieter Abbeel.
\newblock Denoising diffusion probabilistic models.
\newblock {\em CoRR}, abs/2006.11239, 2020.

\bibitem{QualcommStable}
Jilei Hou and Ziad Asghar.
\newblock World's first on-device demonstration of stable diffusion on an
  android phone, 2023.

\bibitem{StyleGAN}
Tero Karras, Samuli Laine, and Timo Aila.
\newblock A style-based generator architecture for generative adversarial
  networks.
\newblock {\em CoRR}, abs/1812.04948, 2018.

\bibitem{VAE}
Diederik~P Kingma and Max Welling.
\newblock Auto-encoding variational bayes, 2022.

\bibitem{Reformer}
Nikita Kitaev, Lukasz Kaiser, and Anselm Levskaya.
\newblock Reformer: The efficient transformer.
\newblock {\em CoRR}, abs/2001.04451, 2020.

\bibitem{WinogradConv}
Andrew Lavin.
\newblock Fast algorithms for convolutional neural networks.
\newblock {\em CoRR}, abs/1509.09308, 2015.

\bibitem{AppleStable}
Atila Orhon, Michael Siracusa, and Aseem Wadhwa.
\newblock Stable diffusion with core ml on apple silicon, 2022.

\bibitem{MemManager}
Yury Pisarchyk and Juhyun Lee.
\newblock Efficient memory management for deep neural net inference.
\newblock {\em CoRR}, abs/2001.03288, 2020.

\bibitem{DCGAN}
Alec Radford, Luke Metz, and Soumith Chintala.
\newblock Unsupervised representation learning with deep convolutional
  generative adversarial networks.
\newblock In Yoshua Bengio and Yann LeCun, editors, {\em 4th International
  Conference on Learning Representations, {ICLR} 2016, San Juan, Puerto Rico,
  May 2-4, 2016, Conference Track Proceedings}, 2016.

\bibitem{CompVis}
Robin Rombach.
\newblock Github: Compvis/stable-diffusion, 2022.

\bibitem{LatentStableDiffusion}
Robin Rombach, Andreas Blattmann, Dominik Lorenz, Patrick Esser, and
  Bj{\"{o}}rn Ommer.
\newblock High-resolution image synthesis with latent diffusion models.
\newblock {\em CoRR}, abs/2112.10752, 2021.

\bibitem{Unet}
Olaf Ronneberger, Philipp Fischer, and Thomas Brox.
\newblock {U-Net}: Convolutional networks for biomedical image segmentation.
\newblock In {\em Medical Image Computing and {Computer-Assisted} Intervention
  -- {MICCAI} 2015}, pages 234--241. Springer International Publishing, 2015.

\bibitem{DreamBooth}
Nataniel Ruiz, Yuanzhen Li, Varun Jampani, Yael Pritch, Michael Rubinstein, and
  Kfir Aberman.
\newblock Dreambooth: Fine tuning text-to-image diffusion models for
  subject-driven generation, 2023.

\bibitem{Attention}
Ashish Vaswani, Noam Shazeer, Niki Parmar, Jakob Uszkoreit, Llion Jones,
  Aidan~N. Gomez, Lukasz Kaiser, and Illia Polosukhin.
\newblock Attention is all you need.
\newblock {\em CoRR}, abs/1706.03762, 2017.

\bibitem{GroupNorm}
Yuxin Wu and Kaiming He.
\newblock Group normalization.
\newblock {\em CoRR}, abs/1803.08494, 2018.

\bibitem{LogSoftmax}
Bo Yuan.
\newblock Efficient hardware architecture of softmax layer in deep neural
  network.
\newblock In {\em 2016 29th IEEE International System-on-Chip Conference
  (SOCC)}, pages 323--326, 2016.

\bibitem{ControlNet}
Lvmin Zhang and Maneesh Agrawala.
\newblock Adding conditional control to text-to-image diffusion models, 2023.

\end{thebibliography}
}

\end{document}